\documentclass[letterpaper, 10 pt, conference]{ieeeconf}  

\IEEEoverridecommandlockouts                              

\usepackage{hyperref}
\usepackage{graphicx}
\usepackage{datetime}
\usepackage{cite}
\usepackage{amsmath}
\usepackage{array}
\usepackage{svg}
\usepackage{siunitx}
\usepackage[caption=false,font=footnotesize]{subfig}
\usepackage{multirow}

\title{\LARGE \bf
Design and evaluation of a multi-finger skin-stretch tactile interface for hand rehabilitation robots
}

\author{Alexandre L. Ratschat$^{1, 2}$, Rubén Martín-Rodríguez$^{1}$, Yasemin Vardar$^{1}$, Gerard M. Ribbers$^{2}$,\\ and Laura Marchal-Crespo$^{1, 2, 3}$
\thanks{*This work was funded by the Dutch Research Council (NWO, VIDI Grant Nr. 18934)}
\thanks{$^{1}$A. L. Ratschat, R. Martín-Rodríguez, Y. Vardar, and L. Marchal-Crespo are with the Department of Cognitive Robotics, Delft University of Technology, Delft, The Netherlands}%
\thanks{$^{2}$A. L. Ratschat, G. M. Ribbers, and L. Marchal-Crespo are with the Department of Rehabilitation Medicine, Erasmus MC, University Medical Center Rotterdam, Rotterdam, The Netherlands}
\thanks{$^{3}$L. Marchal-Crespo is with the ARTORG Center for Biomedical Engineering Research, University of Bern, Switzerland
    {\tt\small Corresponding author: a.l.ratschat@tudelft.nl}}%
}

\begin{document}

\maketitle
\thispagestyle{empty}
\pagestyle{empty}

\begin{abstract}

Object properties perceived through the tactile sense, such as weight, friction, and slip, greatly influence motor control during manipulation tasks. However, the provision of tactile information during robotic training in neurorehabilitation has not been well explored. 
Therefore, we designed and evaluated a tactile interface based on a two-degrees-of-freedom moving platform mounted on a hand rehabilitation robot that provides skin stretch at four fingertips, from the index through the little finger. To accurately control the rendered forces, we included a custom magnetic-based force sensor to control the tactile interface in a closed loop.
 The technical evaluation showed that our custom force sensor achieved measurable shear forces of $\pm$\,8\,N with accuracies of 95.2--98.4\% influenced by hysteresis, viscoelastic creep, and torsional deformation. The tactile interface accurately rendered forces with a step response steady-state accuracy of 97.5--99.4\% and a frequency response in the range of most activities of daily living.
Our sensor showed the highest measurement-range-to-size ratio and comparable accuracy to sensors of its kind. These characteristics enabled the closed-loop force control of the tactile interface for precise rendering of multi-finger two-dimensional skin stretch.
The proposed system is a first step towards more realistic and rich haptic feedback during robotic sensorimotor rehabilitation, potentially improving therapy outcomes.

\end{abstract}

\section{Introduction}

Robotic systems, in combination with virtual reality (VR), can be used in high-intensity movement training in motivating and safe environments~\cite{gassert_rehabilitation_2018, Basalp2021}, incorporating games that resemble activities of daily living (ADLs), such as object manipulation~\cite{levac_learning_2019}. Yet, the ecological validity of these games is limited, which might explain the limited gains in ADLs after robotic practice~\cite{Lo2017}. A lack of tactile information---e.g., object friction, weight, and slip---during robotic training may contribute to this lack of ecological validity~\cite{walker_tactile_2015}, and thus the transfer of gained skills to ADLs~\cite{levac_learning_2019}. The importance of tactile input in motor performance is illustrated in clinical syndromes, such as tactile apraxia~\cite{Binkofski2001} and sensory ataxia~\cite{Grundmann85}.

Tangential skin deformations, commonly called skin stretch, especially arise during object interactions and could contribute to robotic training. Applying skin stretch can be exploited to render object friction, as demonstrated by Provancher and Silvester using a single-degree-of-freedom (DoF) tactor device on a single finger~\cite{provancher_fingerpad_2009}. In addition to friction, Schorr et al. showed that skin stretch could render virtual object weight and stiffness with two wearable three-DoF tactor devices on the user's index finger and thumb~\cite{schorr_fingertip_2017}. Quek et al. provided interaction force and torque information via a three-finger pinch grasp by combining a three-DoF tactor-based tactile interface with a commercial three-DoF grounded kinesthetic feedback device~\cite{Quek2015}.  

A main limitation of most skin stretch tactile interfaces is that they typically render forces by applying a skin displacement computed through an estimate of the finger pad stiffness (e.g.,~\cite{leonardis_3-rsr_2017, Quek2015}). However, considering the large variability in finger pad stiffness under varying pressure levels and across individuals~\cite{wiertlewski_mechanical_2012}, the rendered force accuracy is limited. This is especially relevant since high-fidelity force rendering is required to promote the transfer of the gained skill in the virtual environment to the real world \cite{levac_learning_2019}. Multi-axis force sensors can overcome this limitation as they enable closed-loop force control for accurate lateral force rendering at the fingerpad~\cite{kamikawa_comparison_2018, van_beek_static_2021}. However, their high cost and bulky size limit their use in skin stretch interfaces. 

In this work, we present the collected requirements, design, and evaluation of a novel multi-finger tactile interface that is intended to simulate two-dimensional skin stretch while training ADLs in virtual environments. The interface is designed to be integrated into the PRIDE hand rehabilitation robot developed by Rätz et al.~\cite{ratz_design_2022}, which enables hand setup in a closed position for patients with hand spasticity. We also describe the design and evaluation of a custom low-cost and compact magnetic-based force sensor to render the skin stretch stimuli in a closed-loop fashion. We evaluated our solution by characterizing the sensor's static and dynamic errors, hysteresis, viscoelastic creep, and torsional sensitivity, as well as the tactile interface's step and frequency responses. Finally, we discuss our results in the context of relevant literature. 

\section{Methods}

\subsection{Tactile interface requirements}
\label{sec:requirements}

We designed our tactile interface with the following requirements in mind: i) integration in PRIDE (Fig.~\ref{fig:methods}a) while minimally impeding PRIDE's range of motion (ROM) and hand donning; ii) two-DoF fingerpad deformations in the proximodistal and mediolateral directions of the fingers (from index to little finger) with maximum deformations of 5\,\si{mm}~\cite{nakazawa_characteristics_2000}; iii) maximum renderable forces of 5\,\si{N}; iv) maximum force rendering error of 50\,\si{mN} for forces below 1.5\,\si{N} and 100\,\si{mN} for larger values based on the just-noticeable difference (JND) for force perception~\cite{brodie_sensorimotor_1984}; v) rise time of 0.35\,\si{s} to render object interaction onset~\cite{ johansson_roles_1984}; vi) bandwidth of 10\,\si{Hz}, the highest movement frequency reported during ADLs~\cite{Mann1989}; vii) total material costs under 200\,€ to keep the technology low-cost and accessible to all. 

\subsection{Two-dimensional skin stretch moving platform}

\begin{figure}[t!]
    \centering
    \footnotesize
    \vspace{.1cm}
    \includegraphics[width=\columnwidth]{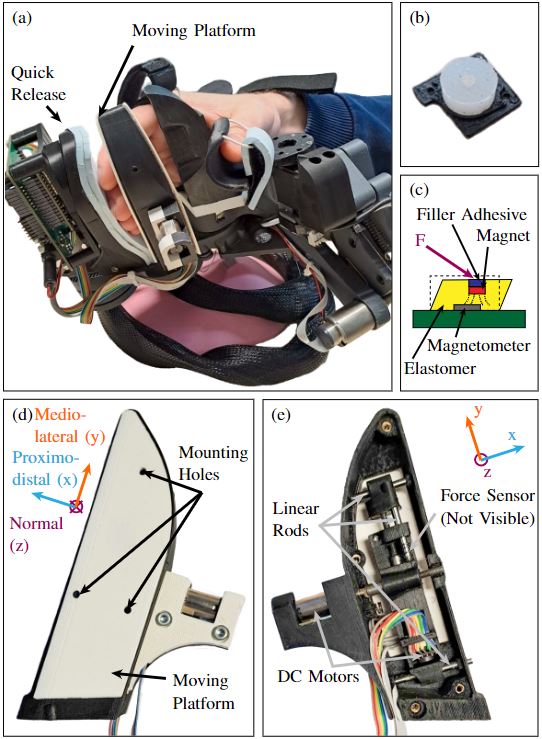} 
    \caption[]{Overview of the tactile interface and custom sensor. (a) The tactile interface is integrated into the PRIDE rehabilitation robot, with the fingerpads touching the moving platform and pressure applied on the fingers' dorsal side with the device's quick release. (b) Custom force sensor with the magnet embedded in the silicone. (c) Schematic showing the sensor elastomer deforming under a force \textbf{F}. (d) Front view of the tactile interface with the moving platform.  
    (e) Back view of the tactile interface with linear rods, DC motors, and force sensor.}
    \label{fig:methods}
    \vspace{-.6cm}
\end{figure}

The skin stretch tactile interface was designed to replace PRIDE's existing finger end-effector (Fig.~\ref{fig:methods}a). We took advantage of the current quick-release finger fixation at the back of the fingers to ensure that the platform movements efficiently transmit into the deformations of the fingerpads. 

The platform (Fig.~\ref{fig:methods}d-e) was actuated with two lead screws (M3x30\,\si{mm} stainless steel threaded rod) in the fingers' proximodistal (X-axis) and the mediolateral (Y-axis) directions connected to two brushed DC motors (5:1 Micro Metal Gearmotor HPCB 6V with Extended Motor Shaft, Pololu, USA). Each axis included 3\,\si{mm} diameter stainless steel linear rods to stabilize the platform movement. To reduce axle friction, miniature linear bearings (C-LMU3, MISUMI Europe, Germany) and Teflon tubing were used for the X and Y axes, respectively. Each axis achieved a total range of $\pm$4.5\,\si{mm}. A set of custom 3D-printed parts made up the device's housing and provided connections for the actuators and the mechanical components. 

Magnetic rotary encoders (Magnetic Encoder Pair Kit for Micro Metal Gearmotors, Pololu, USA) were mounted at the rear shaft of the DC motors to measure the actuators' positions. Considering the gearbox and lead screw reductions, the encoders provided a theoretical resolution of 8.3\,\si{\mu m}. Each motor was controlled using a DRV8876 brushed DC motor driver (Texas Instruments, USA) using differential pulse-width modulation (PWM) in slow-decay mode. An ESP32 Devkit-V1 microcontroller (Espressif, China) integrated and controlled all the components. A custom-made printed circuit board was designed to host the drivers, regulator, microcontroller, and sensor and actuator connectors. 

\subsection{Custom-made magnetic-field-based force sensor}

To create an accurate closed-loop force rendering system, we decided to place a force sensor between the platform in contact with the fingerpads and the actuation mechanism (available volume $\sim$1\,\si{cm^{3}}). We opted for a custom three-DoF magnetic-field-based force sensor due to its simple and cheap manufacturing process and small size (Fig.~\ref{fig:methods}b-c)~\cite{wang_design_2016}.

The sensor relied on a magnet embedded in an elastomer on top of a three-axis magnetometer (MLX90393, Melexis, Belgium) located 3.5\,\si{mm} beneath the magnet (\textit{air gap}) to increase the sensor's sensitivity. When forces are applied, the elastomer deforms and the magnet is displaced, varying the magnetic field at the magnetometer. The magnetic field was generated by a 3.0$\times$1.0\,\si{mm} cylindrical neodymium magnet with N48 remanence (MAGZ-087-P, MagnetPartner, Denmark), equivalent to $\sim$1.4\,\si{T}. The strong remanence helped to improve the sensor signal-to-noise ratio, while its small size facilitated its embedding within the elastomer. Through a calibration procedure (see section~\ref{sec:calibration}), the sensor could measure the applied forces based on the measured variations in the magnetic field. 

The elastomer's shape (cylindrical), size (11\,\si{mm} diameter, 4.5\,\si{mm} height), and material---a silicone with A30 shore hardness (DragonSkin 30, Smooth-On, USA)---were chosen to achieve the stiffness needed to cover the range of desired measurable forces. The sensor was manufactured by mold-casting the elastomer body with a cavity for embedding the magnet, which was sealed using Sil-poxy (Smooth-On, USA). The elastomer was attached to the platform and the magnetometer board using cyanoacrylate adhesive (SUPERGLUE, Loctite, USA). A grooved pattern on both elastomer surfaces enhanced the adhesive binding strength.

The material costs of the tactile interface, including two sensors, were below 150\,€. Although we embedded only one force sensor in the final design, we manufactured two for evaluation purposes. 

\subsection{Control}
\label{sec:control_design}
The custom sensor's magnetometer was interfaced via Serial Protocol Interface (SPI) using a 10\,\si{MHz} clock signal in single-measurement mode with a 1.55\,\si{ms} reading time. The sensor was configured to provide data on the 16 least significant bits of the 19-bit internal analog-to-digital (ADC) converter along the X and Y axes, providing the highest resolution possible. The resolution was set one bit lower along the Z axis, i.e., bits 1--17 of the ADC, due to the strong magnetic field along that axis. 

The ESP32 microcontroller was programmed using FreeRTOS and the control loop for reading sensor data and sending control actions was set to 500\,\si{Hz}. The encoder readings and custom force sensor readings were low-pass filtered in real-time at 10\,\si{Hz} using Euler-backward discretization to avoid high-frequency noise. The interface featured a voltage-driven proportional-integral-derivative (PID) force controller with manually tuned gains ($k_{px}, k_{py}$) = (13.2, 12)(\si{V/N}), ($k_{ix}, k_{iy}$) = (3.6, 12)(\si{V/Ns}), ($k_{dx}, k_{dy}$) = (0.36, 0.72)(\si{Vs/N}), for the X and Y axes, respectively.

\subsection{Experimental evaluation}

We first calibrated two custom magnetic-based force sensors: a standalone sensor and a sensor mounted on the tactile interface. The standalone sensor was employed to fully evaluate and characterize the sensor type, without space restrictions from the moving platform. The sensor embedded in the tactile interface was used to force control and evaluate the system's step and frequency response.

\subsubsection{Experimental setup}

\begin{figure}[h]
    \centering
    \footnotesize
    \vspace{.1cm}
    \includegraphics[width=\columnwidth]{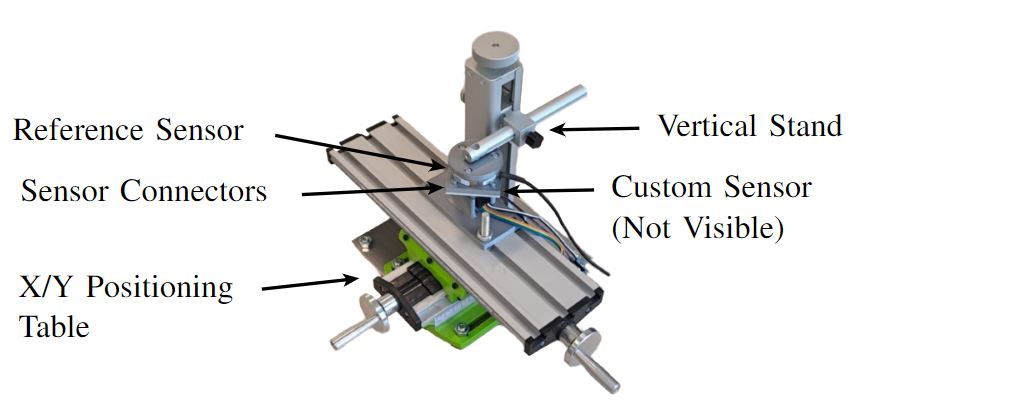}
    \vspace{-.7cm}
    \caption[]{Experimental setup used to calibrate and evaluate the custom sensors and tactile interface.}
    \label{fig:experimental_setup}
    \vspace{-.6cm}
\end{figure}

All measurements and evaluations were conducted with the same experimental setup (Fig.~\ref{fig:experimental_setup}). To measure the forces applied to the sensor and, eventually, exerted by the tactile interface, an ATI Nano 43 force-torque sensor (ATI Industrial Automation, USA) was used (\textit{reference sensor}). A three-axis positioning stage aligned the reference sensor with the standalone custom sensor or the tactile interface (depending on the evaluation) to apply controlled displacements. The stage consisted of a two-axis positioning table (HBM Machines, The Netherlands) connected through a custom mounting plate to a RK-10 stand (Dino-Lite Europe, The Netherlands). The custom sensor or tactile interface was mounted on the positioning table, while the reference sensor was mounted on the stand, allowing relative displacements between the two. Data from the reference sensor were collected via a data acquisition card (USB-6351, National Instruments, USA) and sent to a host computer. The information from the tactile interface, i.e., timestamp, measured forces, desired forces, and commanded control action, was sent to the host computer through the USB port of the ESP32 microcontroller. The management of the communication and the collection of data were performed in Python.

\subsubsection{Sensor calibration}
\label{sec:calibration}

To determine the mapping from magnetic field readings to forces (sensor calibration), we applied controlled deformations to the sensor using the positioning stage and recorded paired magnetic fields--applied force data from the reference sensor, at a sampling rate of 100\,\si{Hz}.

Training and testing data sets were collected to train and test the mapping. The training set consisted of 100 samples collected over 1\,\si{s} at 165 target locations, yielding 16500 samples in total. The target locations were generated as relative displacements between the custom and reference sensor with five equally-distributed steps vertically and radially within [0.0, 0.8]\,\si{mm} and [0.0, -2.0]\,\si{mm}, respectively, and eight angular steps within [0, 360]\,\si{\degree} around the vertical axis. The test set contained 45 randomly generated target locations, yielding 4500 samples in total. 

The magnetic field data were mapped to forces by combining a third-order polynomial expansion with a linear regression---i.e., parameter optimization via stochastic gradient descent with a Huber loss function and L2 regularization term, similar to~\cite{le_signor_gradiometric_2022}. The evaluation of the custom sensor calibration was performed by calculating the \textit{static error} as the L2 norm of the mean absolute errors (MAE) per axis between the reference and custom sensor from the training and test datasets. We also report the \textit{dynamic error}, calculated as the L2 norm of the MAE between the reference and custom sensor readings while varying the applied forces by the positioning stage over 30\,\si{s},  with forces randomly generated within [-5, 4.5]\,\si{N}, [-8, 3]\,\si{N} and [-9, 1]\,\si{N} in the X, Y, and Z axes, respectively. 

\subsubsection{Custom force sensor characterization}

We characterized the custom standalone sensor by evaluating its hysteresis, viscoelastic creep, and torsional sensitivity. To evaluate the potential \textit{hysteric error}, a loading-unloading cycle was applied with the positioning stage to the custom sensor through an indentation of 0.5\,\si{mm} in the Z axis and of 1.25\,\si{mm} in the X and Y axes. During the X and Y measurements, an additional 0.2\,\si{mm} indentation in the Z direction was applied to recreate the working conditions in which the sensor will be used. The hysteresis value at each axis was then computed as the difference in the custom sensor readings at loading and unloading conditions at half of the maximum applied force (measured by the reference sensor) over the maximum applied force (reported in percentage).

To evaluate the \textit{viscoelastic creep}, the same indentations as in the hysteresis test were applied and sustained for a total of 30\,\si{s}. The differences between the readouts of the custom and reference sensors were computed after 5\,\si{s} and 30\,\si{s} of the force onset. The creep was then calculated as the difference between those two values.

To evaluate how the custom sensor readouts were affected by torsional deformations around each axis (\textit{torsional sensitivity}), we incorporated custom connectors between the reference sensor and the positioning stage stand. Each connector aligned with the custom sensor at a 15\,\si{\degree} offset around each of the three axes, i.e., three different connectors were employed. We collected data for 20\,\si{s}, starting with the custom sensor undeformed and ending fully deformed around the respective axis. The L2 norm of the MAE between the reference and standard sensor readouts at rest and fully deformed were then calculated and compared.

\subsubsection{Force control evaluation}

A second force sensor unit was manufactured, mounted within the tactile interface, and calibrated. We performed a step response test and a frequency bandwidth test to evaluate the force-control performance. In both tests, measured forces, desired forces, and the commanded control action were collected at 500\,\si{Hz} from the tactile interface, matching the control frequency of the tactile interface. A normal force of 2\,\si{N} was always applied with the positioning stage.

For the step response test, we recorded the system's response during 10 steps with an amplitude of 1, 2, and 3\,\si{N} along the positive X, Y, and XY directions (30 steps per direction). Each step lasted 2\,\si{s} and was preceded and followed by 1\,\si{s} of zero force. The steady-state error was computed as the error after 1.8\,\si{s} of the step onset, while the rise time was computed as the time required to rise from 10\% to 90\% of the steady-state value of the custom sensor signal. These metrics were computed for each step and averaged across the 30 steps per direction. 

For the bandwidth test, we commanded an exponential chirp signal that swept from 1 to 100\,\si{Hz} at a constant amplitude of 1\,\si{N} (2\,\si{N} peak-to-peak) in the X and Y directions, for 10 seconds each. We applied a Hann window to the commanded forces and the forces measured by the custom sensor to reduce spectral leakage, followed by a Fast Fourier Transform (FFT). The magnitudes and phases of the Bode plots illustrating the system's frequency responses were calculated as the magnitudes and phases of the measured force FFTs divided by the commanded force FFTs. 

\section{Results}

\subsection{Sensor calibration}

During the calibration procedure of the custom standalone sensor, we recorded maximum absolute forces of approximately (8, 8, 16)\,\si{N} in the X, Y, and Z axes, respectively. 
The static error for the training and test datasets of the standalone sensor, as well as the dynamic error, can be found in Table~\ref{tab:accuracy}. 

\begin{table}[ht]
    \centering
    \caption{Standalone custom sensor static and dynamic error}
    \vspace{-.2cm}
    \label{tab:accuracy}
    \resizebox{\columnwidth}{!}{%
    \begin{tabular}{lc}
    \textbf{Evaluation (Dataset)} & \textbf{Error X, Y, Z (MAE $\pm$ SD) / N} \\
    \hline
    Static Error (Training) & $0.068 \pm 0.102$, $0.056 \pm 0.053$, $0.250 \pm 0.260$ \\

    Static Error (Test) & $0.090 \pm 0.072$, $0.152 \pm 0.100$, $0.390 \pm 0.280$ \\
    
    Dynamic Error & $0.108 \pm 0.094$, $0.181 \pm 0.156$, $1.355 \pm 0.627$ \\
    \end{tabular}%
    }
    \vspace{-.5cm}
\end{table}

\subsection{Force sensor characterization}

The hysteresis evaluation of the custom standalone force sensor resulted in values of 11.57\%, 12.39\%, and 13.58\% in the X, Y, and Z axes, respectively. Along the Z-axis, we observed an initial offset of 1.168\,\si{N} under no-load conditions, which increased to 2.717\,\si{N} under the 12.806\,\si{N} maximum load applied during the test. We also observed viscoelastic creep of 0.090\,\si{N}, 0.058\,\si{N}, and 0.508\,\si{N} in X, Y, and Z, respectively. Finally, the measurement errors in the X, Y, and Z directions induced by the torsional distortions were -0.071\,\si{N}, 1.388\,\si{N}, and 4.926\,\si{N} for rotations around the X axis; 1.327\,\si{N}, 0.192\,\si{N}, and 6.538\,\si{N} for the Y axis; and 0.201\,\si{N}, -0.137\,\si{N}, and 0.843\,\si{N} for the Z axis.

\subsection{Force control evaluation}

\subsubsection{Custom embedded sensor}
The calibration of the custom embedded sensor yielded maximum absolute forces of approximately (6, 6, 6)\,\si{N} and test set static errors of ($0.214 \pm 0.109$, $0.095 \pm 0.105$, $1.451 \pm 0.337$)\,\si{N} in the X, Y, and Z axes, respectively. 
\subsubsection{Step response}
We commanded a series of step signals to the interface along different axes. Regarding the steady-state error of the custom sensor w.r.t. the commanded force, average errors of 9\,\si{mN}, 28\,\si{mN}, and 37\,\si{mN} were observed in X, Y, and XY (diagonally), respectively. The rise times, as measured by the custom sensor, were 0.063\,\si{s}, 0.065\,\si{s}, and 0.071\,\si{s} in X, Y, and XY, respectively. 

\subsubsection{Frequency response}

\begin{figure}[h]
    \centering
    \footnotesize
    \includegraphics[width=\columnwidth]{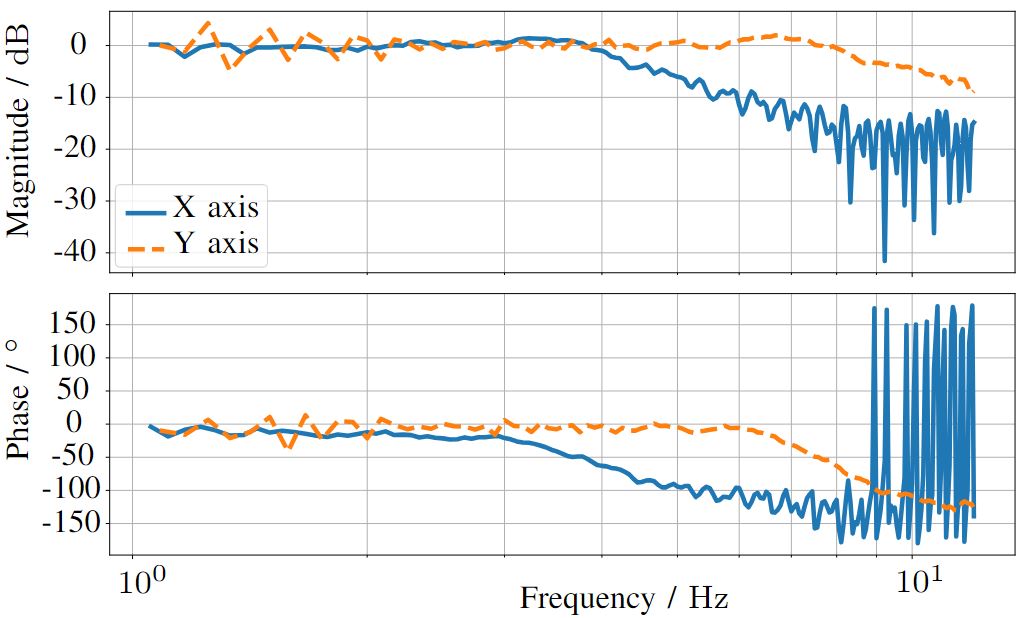}
    \vspace{-.6cm}
    \caption[]{Bode plots of the tactile interface's frequency response for the X and Y axes.}
    \label{fig:bode_plot}
    \vspace{-.4cm}
\end{figure}

We commanded a chirp signal of constant amplitude along each system axis. In the bode plots (Fig.~\ref{fig:bode_plot}), we observed gain-crossover frequencies at approximately 4\,\si{Hz} and 8\,\si{Hz} for the X and Y axes, respectively. 

\section{Discussion}


Despite the importance of tactile information in motor control, its inclusion in robotic neurorehabilitation systems remains largely unexplored. Here, we presented a novel two-dimensional skin stretch tactile interface to extend the haptic rendering capabilities of the PRIDE hand rehabilitation robot to simulate ADLs with high ecological validity.  


\subsection{Compact multi-finger tactile interface providing two-dimensional skin stretch in a hand rehabilitation robot}

Our four-finger tactile interface has the potential to enhance the sensory stimulation capabilities of our hand rehabilitation robot PRIDE by providing skin stretch on top of the already available kinesthetic feedback. This is an advantage w.r.t. current technologies that usually only provide single-finger tactile feedback~\cite{schorr_fingertip_2017, leonardis_3-rsr_2017} and do not enable interactions with tangible virtual objects, such as grasping~\cite{van_beek_static_2021}. 

The actuation mechanism of our interface achieves the required deformations of the fingerpads and maximum renderable forces (5\,\si{mm} and 5\,\si{N}). However, the compliance of the custom force sensor reduced these maximum values to an effective ROM of $\pm$3\,\si{mm} and 3\,\si{N} lateral force, similar to other state-of-the-art devices (e.g.,~\cite{schorr_fingertip_2017}). With the added capabilities, the full range of motion of PRIDE with three DoFs was kept intact, except for small reductions in ROM for the thumb flexion/extension and circumduction (by approximately 15\,\si{\degree}).


\subsection{Small size magnetic-based custom 3D force sensor with large measurement ranges}
\label{sec:discussion_sensor}

Our low-cost custom magnetic-based sensor recorded the largest shear force ranges ($\pm$8\,\si{N}) for a sensor of this type and small size, compared to those reported in literature (e.g.,  $\pm$1\,\si{N}~ in \cite{wang_design_2016, le_signor_gradiometric_2022, dwivedi_design_2018}). Only one magnetic-based sensor was reported to achieve a similar measurement range, yet it is ten times larger than ours~\cite{nie_soft_2017}. The large measurement-range-to-size ratio of our sensor can partly be attributed to the stiffer elastomer selected. Furthermore, our experimental setup allowed the application of forces with reduced slippage. 


The static and dynamic errors of our sensors (1.56--4.81\,\%) were, in general, higher than those found in the literature (e.g., 0.07--2.2\,\% in~\cite{wang_design_2016, le_signor_gradiometric_2022, dwivedi_design_2018}). This might be due to the limited number of data points employed for calibration. Due to the time-consuming manual positioning stage in our calibration procedure, we only recorded samples at 165 target positions for the training of our magnetic-field-to-force mapping, a number substantially smaller than those employed in previous studies that used automated positioning stages, e.g., 4800 in~\cite{wang_design_2016} and 13000 in~\cite{le_signor_gradiometric_2022}.

The observed force offset along the Z axis (approximately 7.3\% of the maximum force) could result from permanent elastomer deformations during calibration and temperature changes affecting the elastomer's stiffness. Le Signor et al. report a 23\% increase in measured force in the Z axis for a temperature increase of 50\,\si{\degree C}~\cite{le_signor_gradiometric_2022}. We chose not to incorporate temperature compensation as it increases the measurement time by 0.25--1.6\,\si{ms}, depending on the oversampling ratio.


We measured a hysteresis of 11--13\,\% in all axes, in contrast to reported values in literature, e.g., 3.4\,\% in~\cite{wang_design_2016} and no hysteresis in~\cite{dwivedi_design_2018} in the normal direction. There is, however, no comparable data from similar sensors in the lateral directions. This difference might originate in the larger deformations applied to our sensor and to the calibration database that only contained measurements collected during static, step-wise loading. 

The measurement of viscoelastic creep in the lateral directions is a novelty for this type of sensor. We observed a force change of approximately 2\,\% of the applied force in the X and Y axes and 5\,\% in the Z-axis. This is in line with the 3\,\% creep measured in the normal direction by Wang et al., despite the difference in materials, i.e., shore 00-30 hardness vs. A30 in our sensor~\cite{wang_design_2016}. Notably, we observed that creep effects are less prominent in the shear vs. the normal direction, likely due to the material properties of the elastomer.


Finally, the applied torsional deformations around the X and Y axes caused errors and drifts in the measured normal force. This result highlights the importance of physically constraining torsional deformations, especially around the X and Y axes, to achieve accurate force readings.


\subsection{Closed-loop force rendering in a compact solution}
\label{sec:discussion_force_control}

The inclusion of the compact and low-cost force sensor allows for closed-loop rendering of lateral forces at the fingertips. This addresses the limitations of current systems that depend on individual calibration of finger pad stiffness for accurate force rendering~\cite{leonardis_3-rsr_2017, schorr_fingertip_2017} and systems requiring costly or bulky force sensors~\cite{van_beek_static_2021}. 

The tactile interface's step-response steady-state errors (0.6--2.5\%) 
indicate accurate force rendering of the system. The measured errors (9--37\,\si{mN}) were smaller than the required maximum errors to be below the JND for force perception (50--100\,\si{mN})~\cite{brodie_sensorimotor_1984}. However, it should be noted that the overall force rendering accuracy of the system is limited by the accuracy of the custom force sensor. Yet, we argue that the system's lateral force rendering quality is an improvement to those solutions relying on fingerpad stiffness estimations. Fingertip stiffnesses can be as much as two times larger from one individual to the other, or for one person between two directions, i.e., mediolateral vs. proximodistal~\cite{wiertlewski_mechanical_2012}. Thus, our solution has the potential to eliminate the need for tedious individual fingerpad stiffness calibration procedures. 

The observed rise times (63--71\si{ms}) were smaller than those measured for lifting onset during object manipulation (approximately 350\,\si{ms}~\cite{johansson_roles_1984}), fulfilling our requirement. However, the effective bandwidth of the interface (4--8\,\si{Hz} gain crossover) was slightly lower than the targeted 10\,\si{Hz} and the effective bandwidth of comparable skin stretch devices (e.g., 10-12\,\si{Hz} gain crossover~\cite{kamikawa_comparison_2018}). This might be due to compliance added from the deformable force sensor, actuation backlash, and the filtering in the control strategy. The compliance of the force sensor acts as a mechanical low-pass filter, limiting the effective bandwidth of the system and adding lag. While the frequency response of the system is sufficient to render a wide range of object interactions during ADLs~\cite{Mann1989}, especially during neurorehabilitation, a higher bandwidth would allow for accurate rendering of extreme cases, such as collisions. 

\subsection{Limitations \& Future Work}

Most of the limitations of our tactile interface, such as the low accuracy of the custom sensor and the limited frequency response, stem from our requirement to create a low-cost and accessible system. Investing in high-quality force sensors and actuators may result in a more accurate and stiff system, and alleviate the need for lengthy calibration procedures. 

The employed control strategy was a first effort to show the capabilities of the developed system but still lacked refinement regarding filtering, parameter tuning, and system modeling. Future work should contribute towards increasing the system's force rendering quality, i.e., the force rendering accuracy and bandwidth, by improving or exchanging the sensor, actuation mechanism, and control strategy. Finally, the effect of training with the combined haptic feedback of PRIDE and the tactile interface on motor control should be investigated with people with motor and sensory deficits. 

\section{Conclusion}

Existing robotic neurorehabilitation systems largely fail to provide tactile information, a crucial component for motor control when interacting and manipulating objects. We presented a novel low-cost tactile interface to provide skin stretch stimulation in the mediolateral and proximodistal directions of the little to index fingers to extend the capabilities of hand rehabilitation robots to render interactions with virtual tangible objects as a means to enhance the ecological validity of robotic neurorehabilitation training. We manufactured a custom low-cost three-axis magnetic-based force sensor with the highest measurement-range-to-size ratio of its kind to achieve a closed-loop force control scheme. The forces rendered by the tactile interface showed low steady-state errors and a frequency response sufficient to simulate a wide range of activities of daily living. We hope that this study serves as a stepping stone towards including tactile rendering in robotic neurorehabilitation and, ultimately, enhance the recovery and quality of life for people after a stroke.




\bibliographystyle{IEEEtran}
\bibliography{references}

\addtolength{\textheight}{-12cm}   

\end{document}